\documentclass{article}

    \PassOptionsToPackage{numbers, compress}{natbib}

\usepackage{graphicx} 
 \usepackage{amsmath}
 \usepackage{multirow} 
 \usepackage{listings}
\usepackage{float}
\usepackage{wrapfig}
    \usepackage[preprint]{neurips_2024}

\usepackage[utf8]{inputenc} 
\usepackage[T1]{fontenc}    
\usepackage{hyperref}       
\usepackage{url}            
\usepackage{booktabs}       
\usepackage{amsfonts}       
\usepackage{nicefrac}       
\usepackage{microtype}      
\usepackage{xcolor}         
\usepackage{tcolorbox}

\title{Leveraging 2D Information for Long-term Time Series Forecasting with Vanilla Transformers}

\author{Xin Cheng$^{1*}$ \quad   Xiuying Chen$^{2*}$ \quad Shuqi Li$^{3}$ \quad Di Luo$^{3}$\quad   \\\bf Xun Wang$^{4}$ \quad Dongyan Zhao$^{1}$ \quad Rui Yan$^{3}$ \\
\\
$^1$\ Peking University \quad $^2$\ KAUST \quad $^3$\ Renmin University of China \quad $^3$\ Microsoft\\ 
chengxin1998@stu.pku.edu.cn \quad xiuying.chen@kaust.edu.sa}

\begin{document}

\maketitle

\begin{abstract}
 Time series prediction is crucial for understanding and forecasting complex dynamics in various domains, ranging from finance and economics to climate and healthcare.
Based on Transformer architecture, one approach involves encoding multiple variables from the same timestamp into a single temporal token to model global dependencies. 
In contrast, another approach embeds the time points of individual series into separate variate tokens. 
The former method faces challenges in learning variate-centric representations, while the latter risks missing essential temporal information critical for accurate forecasting.
In our work, we introduce GridTST, a model that combines the benefits of two approaches using innovative multi-directional attentions based on a vanilla Transformer.
We regard the input time series data as a grid, where the $x$-axis represents the time steps and the $y$-axis represents the variates.
A vertical slicing of this grid combines the variates at each time step into a \textit{time token}, while a horizontal slicing embeds the individual series across all time steps into a \textit{variate token}.
Correspondingly, a \textit{horizontal attention mechanism} focuses on time tokens to comprehend the correlations between data at various time steps, while a \textit{vertical}, variate-aware \textit{attention} is employed to grasp multivariate correlations.
This combination enables efficient processing of information across both time and variate dimensions, thereby enhancing the model's analytical strength.
The GridTST model consistently delivers state-of-the-art performance across various real-world datasets. 
We further conduct a comprehensive analysis to showcase the effectiveness of different components of our model, its generalization capability across diverse variates and time series, and its enhanced proficiency in utilizing arbitrary lookback windows.
We will release our code and checkpoints to facilitate further research\footnote{\url{https://github.com/Hannibal046/GridTST}}.
\end{abstract}

\section{Introduction}
\begin{figure*}[ht]
\centering
\includegraphics[width=13cm]{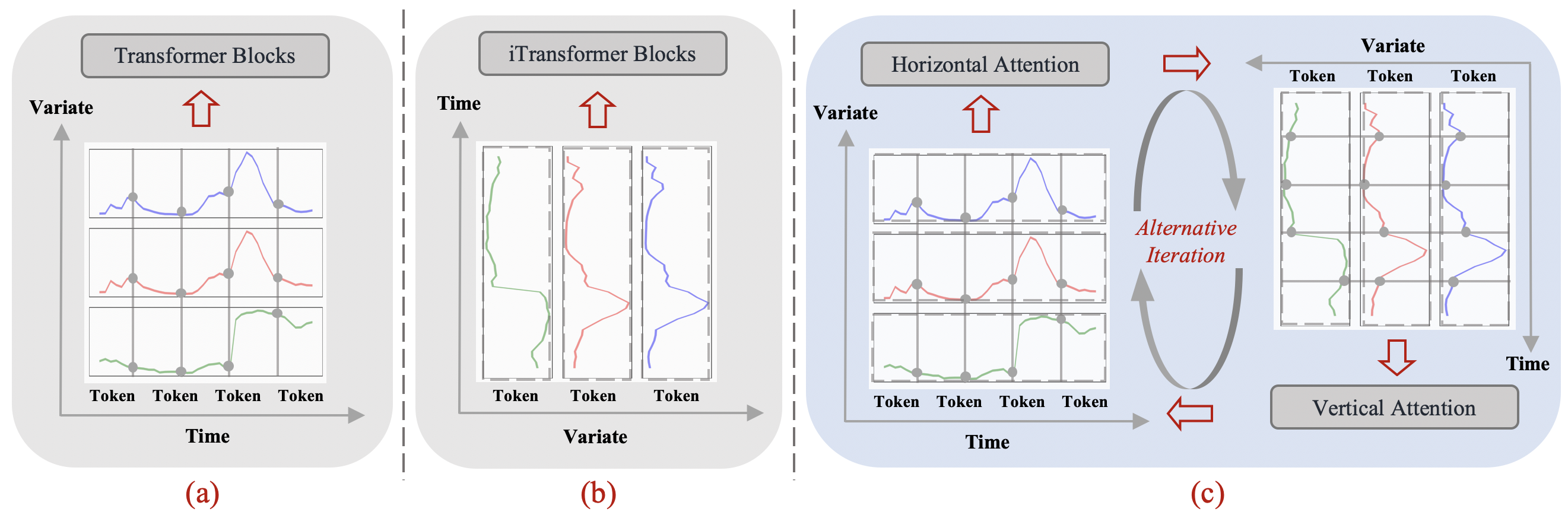}
\caption{Comparison of the vanilla Transformer (a), inverse Transformer (b), and our proposed GridTST (c). Unlike baseline transformers that embed time steps into temporal tokens or the entire series into variate tokens separately, GridTST models both simultaneously. This approach captures multivariate and multi-time-step correlations using a bi-directional attention mechanism.}
\label{fig:intro}
\end{figure*}

Time series prediction involves analyzing and interpreting data points collected or recorded at successive time intervals to forecast future values~\cite{li2019enhancing,ismail2019deep}.
This task is fundamental in various fields, from finance and economics to weather forecasting and inventory management, where understanding trends, patterns, and potential future occurrences based on historical data is crucial.
Among all the deep models, there is particular interest in applying Transformer architecture in time series prediction due to its strong performance, computational efficiency, and scalability~\cite{borovykh2017conditional,ding2019modeling,wen2022transformers,hartvigsen2022detecting,liu2023itransformer}. 

The Transformer architecture, introduced in~\cite{vaswani2017attention}, is a groundbreaking neural network design that has transformed how we handle sequential data in natural language processing~\cite{floridi2020gpt,cheng2023lift,chen2023follow} and computer vision~\cite{dosovitskiy2020image,li2020vmsmo}. 
When adapted for time series forecasting, the Transformer utilizes its ability to handle sequential dependencies and complex patterns, making it highly effective in predicting future trends and values in time-dependent data~\cite{ding2019modeling,agarwal2022real,zhou2022fedformer,melkas2021interactive}.

Classic transformer-based forecasters embed multi-variate data points from the same timestamp into a single variable and use temporal tokens to capture temporal dependencies, as illustrated in Figure~\ref{fig:intro}(a). \cite{liu2023itransformer} noted that single-time-step tokens may not effectively convey information due to their limited receptive field. Furthermore, \cite{zeng2023transformers} pointed out the inappropriate use of permutation-invariant attention in the temporal dimension. To address these issues, \cite{liu2023itransformer} developed the iTransformer, which inverts the roles of the attention mechanism and feed-forward network, representing time points as variate tokens (Figure~\ref{fig:intro}(b)). However, this model still lacks adequate timestamp modeling. Effective time modeling is vital in time series prediction to capture unique patterns like trends and seasonality, as emphasized by \cite{zhou2022fedformer,nie2022time,kiritoshi2021estimating}. These models employ the vanilla Transformer architecture, which ensures generalizability and compatibility with existing hardware accelerators, aiding in deployment and scalability for complex datasets. This raises a critical question: Can the vanilla Transformer architecture effectively capture both temporal and covariate information?

To address the above issues, in this work, we propose a novel method named GridTST, which captures the cross-time and cross-variate dependency by adapting the traditional attention mechanism and architecture from different views.
Specifically, we adapt the Patching technique from \cite{nie2022time} into our mechanism. 
This involves segmenting time series data into subseries-level patches that function as input tokens. 
The motivation behind this is that a single time step, unlike a word in a sentence, lacks inherent semantic meaning. 
Therefore, extracting local semantic information becomes crucial for analyzing their interconnections.
Then, we visualize the input time series data as a grid, with the $x$-axis representing time steps and the $y$-axis representing the variates. 
By slicing this grid vertically, we combine variates at each time step into a `time token', and through horizontal slicing, we embed the data of individual series across all time steps into a `variate token'. 
Correspondingly, we can apply a horizontal attention mechanism on time tokens to analyze correlations between data at different time steps, and a vertical, variate-aware attention to capture the multivariate correlations. 
In this way,  multivariate and multi-time-step correlations can be depicted by the attention mechanism.
This method is illustrated in Figure~\ref{fig:intro}(c).
Correspondingly, the feed-forward network is able to learn generalizable representations that encapsulate both series- and time-aware features.
The comparison with the existing method frameworks is shown in Table~\ref{compare} in Appendix.

Our contributions lie in three aspects:
Firstly, our observations confirm that both temporal and covariate information are crucial for the task of time series prediction. 
Secondly, we introduce GridTST, a model that effectively leverages the foundational Transformer architecture. It employs an innovative data structuring technique to concurrently and efficiently capture both temporal dynamics and covariate information.
Finally, GridSTS achieves consistent state-of-the-art on real-world forecasting benchmarks. 

\section{Related Work}


In recent years, a multitude of deep models have been introduced for temporal modeling. Recurrent Neural Networks (RNNs) are commonly utilized to capture temporal dependencies, as noted by \cite{yu2017long,salinas2020deepar}. \cite{lai2018modeling} introduced LSTNet, which integrates Convolutional Neural Networks (CNNs) with recurrent-skip connections. Moreover, a body of research has focused on leveraging Temporal Convolutional Networks (TCNs) to understand temporal causality through causal convolution, as exemplified in seminal works by \cite{oord2016wavenet,sen2019think,borovykh2017conditional,bai2018empirical}. Additionally, recent studies such as those by \cite{zeng2022transformers} and \cite{chen2023tsmixer} have explored the use of simpler linear layers for the modeling of complex temporal data, resulting in the development of models like DLinear and TSMixer, respectively.

There are also a large body of literature introducing Transformer~\cite{vaswani2017attention} into time series analysis with specific modules designed for the unique aspects of time series, such as decomposition and periodicity detection. Decomposition stands as a foundational technique that breaks down complex time series into simpler, more predictable components~\cite{wu2021autoformer, zhou2022fedformer, woo2022etsformer}. Inspired by stochastic process theory, the aspect of periodicity within time series is increasingly being factored into models to better handle complex temporal variations~\cite{wu2021autoformer, zhou2022fedformer, woo2022etsformer, wu2022timesnet}. As one of these, TimesNet~\cite{wu2022timesnet} explores the multi-periodicity of time series and captures the intra-period variations and inter-period variations of a single variate simultaneously which is also the first task-general foundation model achieving significant results across all five time series analysis tasks. Besides, Crossformer~\cite{zhang2022crossformer} models the cross-time dependency and the cross-dimension dependency to bridge the gap in focus on modeling the relationships among multivariate data. 

The research landscape for adapting the Transformer architecture to time series analysis includes a promising line of inquiry that prioritizes zero modification of the original Transformer design. This approach ensures generalizability, compatibility with existing hardware accelerators, and scalability—all vital for efficient deployment. Two significant contributions in this domain are PatchTST~\cite{nie2022time} and iTransformer~\cite{liu2023itransformer}. PatchTST introduces channel-independence and patching strategies for effective time-dependency modeling, which have demonstrated consistent performance enhancements and have quickly gained traction in subsequent studies~\cite{jin2023time,zhou2023one}. Contrasting with earlier methods that focus on global dependencies across temporal tokens in time series, iTransformer presents a novel perspective by treating individual series as variate tokens. This approach enables the model to capture intricate multivariate correlations. In the same vein of leveraging the unaltered Transformer architecture, GridTST emerges as a novel model by viewing the time series as a grid and modeling with alternated self-attention, thereby becoming the first to concurrently harness both temporal and covariate information using the vanilla Transformer framework.

\section{GridTST}

\subsection{Problem Formulation}

In multivariate time series forecasting, we take the historical observations as input, denoted by $X = \{x_1, \ldots, x_T\} \in \mathbb{R}^{T \times N}$, where $T$ represents the number of time steps and $N$ the number of variates. 
Our objective is to forecast $F$ future time steps, which we denote as $Y = \{x_{T+1}, \ldots, x_{T+F}\} \in \mathbb{R}^{F \times N}$. 
We denote the prediction results as $\hat{Y} = \{\hat{x}_{T+1}, \ldots, \hat{x}_{T+F}\} \in \mathbb{R}^{F \times N}$. 
For brevity, $X_{t,:}$ refers to the multivariate data recorded at a single time step $t$, whereas $X_{:,n}$ represents the entire time series for a specific variate $n$.

 \subsection{Model Structure}
 
\begin{figure*}[ht]
\centering
\includegraphics[width=13cm]{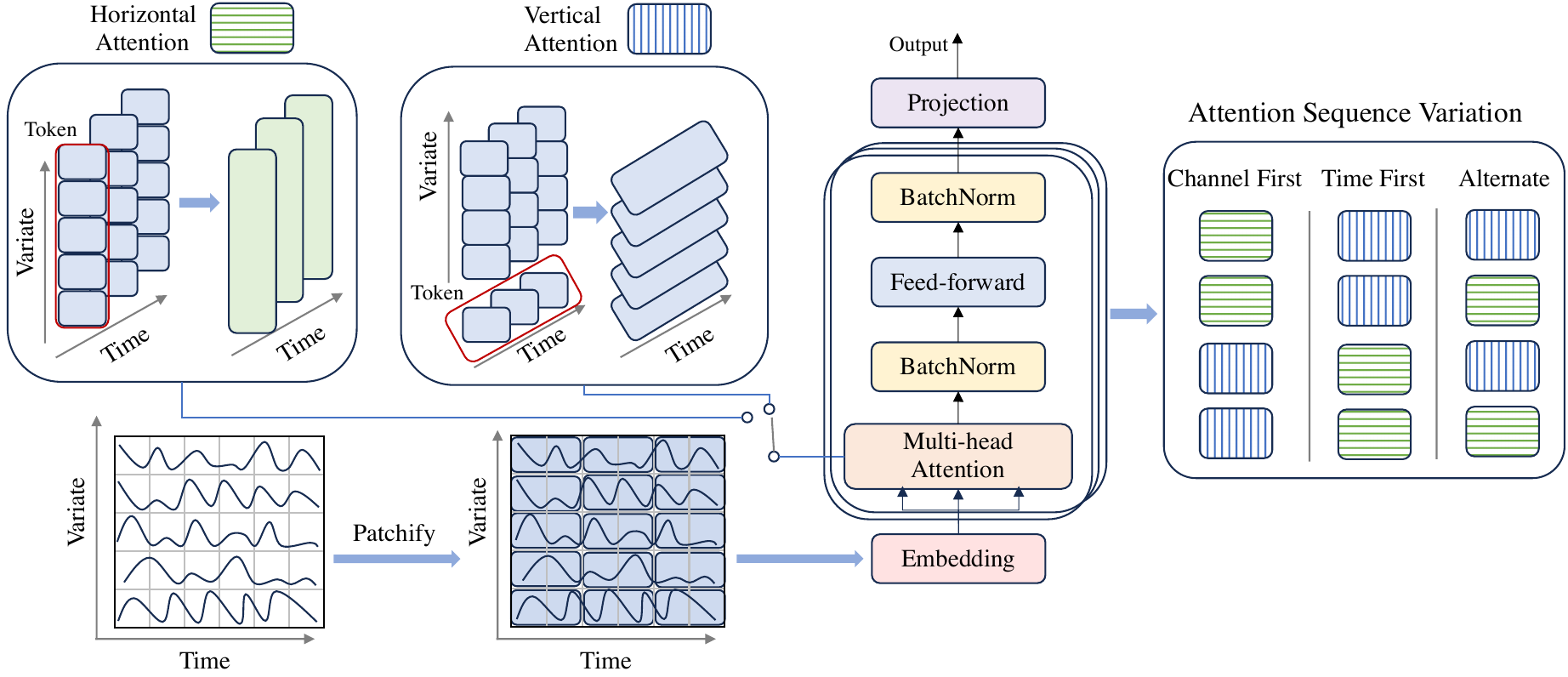}
\caption{Overview of our proposed GridTST. 
Firstly, we transform the inputs by breaking them down into grids. 
These grids undergo processing via vanilla transformer attention, incorporating distinct horizontal and vertical directions. 
Finally, our model generates projected prediction results.}
\label{fig:model}
\end{figure*}

As illustrated in Figure~\ref{fig:model}, GridTST utilizes the vanilla encoder-only architecture of Transformer, including the patched embedding, horizontal and vertical attentions.
It's important to note that the horizontal and vertical attentions are standard Transformer attention modules, with the innovation lying in the dual-view of time series data.

\textbf{Patched Time Tokens.} 
A straightforward approach to generate input tokens involves considering each variate at every time step as a distinct token. 
Nevertheless, in contrast to words within a sentence, an isolated time step does not convey semantic value, underscoring the importance of extracting local semantic features to effectively analyze their interconnections. 
Consequently, we enhance the detection of subtle semantic nuances, which may elude analysis at the individual point level, by aggregating several consecutive time steps into segments, thus creating tokens at the subseries level, following PatchTST~\cite{nie2022time}.

Specifically, we take each univariate input time series $X_{:,n} \in \mathbb{R}^{T \times 1}$ and segment it into patches, which may overlap or be distinctly separate.
The length of each patch is defined as $P$, while the stride $S$ is the distance between the start of one patch and the start of the next. 
This segmentation results in a series of patches $X^p_{:,n} \in \mathbb{R}^{M \times P}$, with the total number of patches given by $M = \left\lceil \frac{T-P}{S} \right\rceil + 2$. 
To maintain continuity at the boundary, we append $S$ copies of the final value $X_{T,n} \in \mathbb{R}$ to the sequence before initiating the patching process.

With the use of patches, the number of input tokens can be reduced from $T$ to approximately $T/S$. 
This implies the memory usage and computational complexity of the attention map are quadratically decreased by a factor of $S$. 
Thus constrained on the training time and GPU memory, patch design can allow the model to see the longer historical sequence.

Following, the patches are projected into the Transformer's latent space of dimension $D$ using a trainable linear projection $W_p \in \mathbb{R}^{P \times D}$, supplemented by a learnable additive position encoding $W_{pos} \in \mathbb{R}^{M \times D}$, to maintain the temporal order of patches: $X^d_{:,n} =  X^p_{:,n}W_p + W_{pos}$, where $X^d_{:,n} \in \mathbb{R}^{M \times D}$ represents the inputs for the Transformer encoder.

Henceforth, we define our input grid as $X^d=\{X^d_1,...,X^d_{M}\} \in \mathbb{R}^{M\times N \times D}$. 
Here, $X^d_{t,:}\in \mathbb{R}^{N \times D}$ denotes the multivariate data encapsulated within the patch at step $t$, while $X^d_{:,n}\in \mathbb{R}^{M \times D}$ captures the complete patched time series corresponding to the specific variate.

\textbf{Horizontal Attention}. 
Horizontal attention in the GridTST model plays a crucial role in understanding the time-dependent aspects of time series data. 
It helps the model to analyze how each data point relates to its predecessors and successors over time. 
This mechanism is key for identifying and learning from patterns, trends, and changes that occur throughout the time series, thereby enabling more accurate and informed predictions about future data points. 
Essentially, horizontal attention ensures the model captures the sequential nature and temporal dynamics in time series data.

Concretely, take the $n$-th variate series as an example, we utilize horizontal attention on the patched time tokens $X^d_{:,n}\in \mathbb{R}^{M \times D}$ to model time dependencies.
In the multi-head attention mechanism, each head $h = \{1, \ldots, H\}$ transforms these inputs into query matrices $Q^h_{:,n} = X^d_{:,n} W_h^Q$, key matrices $K^h_{:,n} = X^d_{:,n} W_h^K$, and value matrices $V^h_{:,n} = X^d_{:,n} W_h^V$, where $W_h^Q, W_h^K \in \mathbb{R}^{D \times d_k}$ and $W_h^V \in \mathbb{R}^{D \times D}$. The attention output $O^d_{:,n}\in \mathbb{R}^{M \times D}$ is then obtained using a scaled product:
\begin{align}
    O^d_{:,n} &= \text{Attention}(Q^h_{:,n}, K^h_{:,n},V^h_{:,n})=\text{Softmax}\left(\frac{Q^h_{:,n} (K^h_{:,n})^T}{\sqrt{d_k}}\right) V^h_{:,n}.
\end{align}
Next, $O^d_{:,n}$ will be processed through BatchNorm layers and a feed-forward network with residual connections, as depicted in Figure~\ref{fig:model}.
The overall process is summarized as:
\begin{align}
    O^{d,l}_{:,n}=\text{Attn}_{\text{horizontal}}(O^{d,l-1}_{:,n}),
\end{align}
where $l$ denotes the layer index.

\textbf{Vertical Attention.}
While horizontal attention focuses on understanding patterns over time, vertical attention is designed to capture the relationships between different variates at a given time step. 
This is crucial for accurately modeling and forecasting in scenarios where the interplay between different variables significantly influences the outcomes.
The use of vertical attention also allows the model to overcome the limitations of methods that struggle with learning representations centered around variates. 
It ensures that the GridTST model doesn't overlook the critical inter-variable dynamics that are essential for precise forecasting in multivariate scenarios.

Specifically, vertical attention focuses on the variate tokens $X^d_{t,:} \in \mathbb{R}^{N \times D}$. 
Similar to horizontal attention, in this vertical multi-head attention mechanism, each head processes these inputs into query matrices $\hat{Q}^h_{t,:} = X^d_{t,:} W_h^{\hat{Q}}$, key matrices $\hat{K}^h_{t,:} = X^d_{t,:} W_h^{\hat{K}}$, and value matrices $\hat{V}^h_{t,:} = X^d_{t,:} W_h^{\hat{V}}$. 
Here, $W_h^{\hat{Q}}, W_h^{\hat{K}} \in \mathbb{R}^{D \times d_k}$ and $W_h^{\hat{V}} \in \mathbb{R}^{D \times D}$. The attention output $\hat{O}^d_{t,:}\in \mathbb{R}^{N \times D}$ is subsequently derived through a scaled product:
\begin{align}
    \hat{O}^d_{t,:} &= \text{Attention}(\hat{Q}^h_{t,:}, \hat{K}^h_{t,:},\hat{V}^h_{t,:})=\text{Softmax}\left(\frac{\hat{Q}^h_{t,:} (\hat{K}^h_{t,:})^T}{\sqrt{d_k}}\right) \hat{V}^h_{t,:}.
\end{align}
This multi-head attention block also incorporates BatchNorm layers and a feed-forward network with residual connections. 
Similarly, the process is denoted as:
\begin{align}
\hat{O}^{d,l}_{:,n}=\text{Attn}_{\text{vertical}}(\hat{O}^{d,l-1}_{t:}).
\end{align}

The core code is in Figure~\ref{code} in Appendix.

\textbf{Attention Sequencing.}
In our GridSTS model, the order of applying horizontal and vertical attentions can be flexibly configured to optimize performance. We explored three distinct configurations:
1. Time-first: Applying horizontal attention first, followed by vertical attention.
2. Channel-first: Applying vertical attention first, followed by horizontal attention.
3. Alternate: An iterative approach where vertical and horizontal attentions are applied in an alternating manner.

Our experimental results in \S~\ref{ablation} demonstrate the effectiveness of these different configurations. 
Notably, we discovered that the sequence starting with vertical attention and then transitioning to horizontal yields the best performance.
This is likely because vertical attention first captures complex variate relationships, laying the groundwork for horizontal attention to then effectively grasp temporal patterns, resulting in more accurate forecasts.

\textbf{Complexity} To offer a clear perspective, consider a time series with $m$ covariates and $n$ patches, and a Transformer model with a hidden size of $d$. The computational complexity of the attention layer for PatchTST is $\mathcal{O}(n^2d)$, whereas for GridTST, it is $\mathcal{O}\left(\frac{m^2d}{2} + \frac{n^2d}{2}\right)$. This implies that for datasets with a relatively small number of covariates---such as ETTh1 and ETTm1, which have seven covariates---GridTST can operate more efficiently than PatchTST. For dataset with large covariate number, we also design an efficient training algorithm by variate sampleing, as detailed in \S~\ref{section:efficient_training_strategy}.

\textbf{Normalization.} 
We denote the representation after the attention sequence process as $Z \in \mathbb{R}^{M \times N \times D}$. 
This representation is subsequently processed through a flatten layer with a linear head to generate the prediction results. 
Notably, we apply instance normalization before patching and instance denormalization after this linear head. 
This approach, recently introduced, effectively mitigates the distribution shift between training and testing data~\cite{ulyanov2016instance,kim2021reversible}. 
Specifically, it normalizes each variate series instance $X_{t,:}$ to have zero mean and unit standard deviation. 
Essentially, each $X_{t,:}$ is normalized before patching, and then its mean and deviation are reintegrated into the output prediction. 
The final output, denoted as $\hat{Y} = \{\hat{x}_{T+1}, \ldots, \hat{x}_{T+F}\} \in \mathbb{R}^{F \times N}$, represents the forecast over the horizon $F$.

\textbf{Loss Function.} 
We employ the Mean Squared Error (MSE) loss to quantify the difference between our predictions and the actual ground truth. 
The loss for each channel is accumulated and then averaged across $F$ time series to calculate the overall objective loss, expressed as $\mathcal{L} = \left\| \hat{X}_{L+1:L+F} - X_{L+1:L+T} \right\|_2^2$.



\section{EXPERIMENTS}


 \textbf{Datasets.} We conduct experiments on seven real-world datasets to evaluate the performance of the proposed GridTST. The first four datasets are collected by \cite{wu2021autoformer}: \textit{Weather} includes 21 meteorological factors collected every 10 minutes from the Weather Station of the Max Planck Biogeochemistry Institute in 2020; \textit{Traffic} collects hourly road occupancy rates measured by 862 sensors of the San Francisco Bay area freeways from January 2015 to December 2016; \textit{Electricity} records the hourly electricity consumption data of 321 clients; \textit{Illness} dataset collects the number of patients and influenza-like illness ratio on a weekly frequency. 
 Additionally, \textit{ETTh1} and \textit{Ettm1} by \cite{chen2023tsmixer} contain 7 factors of electricity transformer from July 2016 to July 2018, recorded hourly. 
 Finally, \textit{Solar-Energy} by \cite{lai2018modeling} records the solar power production of 137 PV plants in 2006, sampled every 10 minutes.
 Details are in Appendix~\ref{appendix_data}.

To ensure consistency in our evaluation, we adhere to the data processing protocol and forecasting length settings as established in TimesNet \cite{wu2022timesnet} and SCINet \cite{liu2022scinet}. 
Specifically, for the Weather, Traffic, Electricity, Ett, and Solar-Energy datasets, we maintain a fixed length of 96 for the lookback series. 
The prediction length, however, varies and is set within the range of \{96, 192, 336, 720\}, allowing us to assess the model's performance across different forecast horizons.

 \subsection{Prediction Performance}

\textbf{Baselines and Experimental Settings.} 
Our study incorporates state-of-the-art Transformer-based models as baseline comparisons. These models include iTransformer~\cite{liu2023itransformer}, PatchTST~\cite{nie2022time}, Autoformer~\cite{wu2021autoformer}, CrossFormer~\cite{zhang2022crossformer}, Informer~\cite{zhou2021informer}, Pyraformer~\cite{liu2021pyraformer}, along with a recent non-Transformer-based model, DLinear~\cite{zeng2023transformers}. Each of these models adheres to a uniform experimental setup. Specifically, for the Illness dataset, the prediction length \(T\) is set within the range \{24, 36, 48, 60\}, while for other datasets, \(T\) spans \{96, 192, 336, 720\}, consistent with the methodologies detailed in their respective original papers. Notably, all models utilize a default look-back window \(L = 336\), in alignment with the DLinear model's parameters. Our evaluation criteria focus on the Mean Squared Error (MSE) and Mean Absolute Error (MAE) in the context of multivariate time series forecasting.

\begin{table*}[t]
	\centering
 	\caption{Multivariate long-term forecasting results with GridTST. We use prediction lengths $T\in \{24, 36, 48, 60\}$ for Illness dataset and $T\in \{96, 192, 336, 720\}$ for the others. The best results are in \textbf{\textcolor{red}{red}} and second best is in \textbf{\textcolor{blue}{blue}}.
 We fix the lookback length T = 336.}
	\resizebox{1.03\linewidth}{!}{
		\begin{tabular}{cc|c|cc|cc|cc|cc|cc|cc|cc|ccc}
			\toprule
& \multicolumn{2}{c|}{Models} & \multicolumn{2}{c|}{GridTST} & \multicolumn{2}{c|}{iTransformer} & \multicolumn{2}{c|}{PatchTST} & \multicolumn{2}{c|}{DLinear} & \multicolumn{2}{c|}{CrossFormer} & \multicolumn{2}{c|}{Autoformer} & \multicolumn{2}{c|}{Informer} & \multicolumn{2}{c}{Pyraformer} \\
			\cline{1-19}
			&\multicolumn{2}{c|}{Metric}&MSE&MAE&MSE&MAE&MSE&MAE&MSE&MAE&MSE&MAE&MSE&MAE&MSE&MAE&MSE&MAE\\
			\cline{1-19}
		&\multirow{4}*{\rotatebox{90}{Weather}}& 96    & \textcolor{red}{\textbf{0.145}} & \textcolor{red}{\textbf{0.195}} & 0.163 & 0.212 & \textcolor{blue}{0.151} & \textcolor{blue}{0.200} & 0.176 & 0.237 & 0.226 & 0.301 & 0.249 & 0.329 & 0.354 & 0.405 & 0.896 & 0.556 \\
&\multicolumn{1}{c|}{}& 192   & \textcolor{red}{\textbf{0.191}} & \textcolor{red}{\textbf{0.241}} & 0.203 & \textcolor{blue}{0.249} & \textcolor{blue}{0.195} & \textcolor{red}{0.241} & 0.220 & 0.282 & 0.215 & 0.289 & 0.325 & 0.370 & 0.419 & 0.434 & 0.622 & 0.624 \\
&\multicolumn{1}{c|}{}& 336 & \textcolor{red}{\textbf{0.243}} & \textcolor{red}{\textbf{0.280}} & 0.255 & 0.289 & \textcolor{blue}{0.247} & \textcolor{blue}{0.281} & 0.265 & 0.319 & 0.319 & 0.317 & 0.351 & 0.391 & 0.583 & 0.543 & 0.739 & 0.753 \\
&\multicolumn{1}{c|}{}& 720 & \textcolor{red}{\textbf{0.316}} & \textcolor{red}{\textbf{0.333}} & 0.326 & \textcolor{blue}{0.337} & \textcolor{blue}{0.321} & \textcolor{red}{0.333} & 0.323 & 0.362 & 0.381 & 0.379 & 0.415 & 0.426 & 0.916 & 0.705 & 1.004 & 0.934 \\

\cline{1-19}
&\multirow{4}*{\rotatebox{90}{Traffic}}& 96    & \textcolor{red}{\textbf{0.337}} & \textcolor{red}{\textbf{0.241}} & \textcolor{blue}{0.358} & 0.258 & 0.366 & \textcolor{blue}{0.251} & 0.410 & 0.282 & 0.556 & 0.347 & 0.597 & 0.371 & 0.733 & 0.410 & 2.085 & 0.468 \\
&\multicolumn{1}{c|}{}& 192   & \textcolor{red}{\textbf{0.373}} & \textcolor{red}{\textbf{0.259}} & \textcolor{blue}{0.375} & 0.268 & 0.388 & \textcolor{blue}{0.263} & 0.423 & 0.287 & 0.588 & 0.320 & 0.607 & 0.382 & 0.777 & 0.435 & 0.867 & 0.467  \\
&\multicolumn{1}{c|}{}& 336   & \textcolor{red}{\textbf{0.378}} & \textcolor{red}{\textbf{0.259}} & \textcolor{blue}{0.389} & 0.274 & 0.398 & \textcolor{blue}{0.265} & 0.436 & 0.296 & 0.601 & 0.347 & 0.623 & 0.387 & 0.776 & 0.434 & 0.869 & 0.469  \\
&\multicolumn{1}{c|}{}& 720   & \textcolor{red}{\textbf{0.403}} & \textcolor{red}{\textbf{0.276}} & \textcolor{blue}{0.422} & 0.290 & 0.434 & \textcolor{blue}{0.287} & 0.466 & 0.315 & 0.605 & 0.345 & 0.639 & 0.395 & 0.827 & 0.466 & 0.881 & 0.473 \\

\cline{1-19}
&\multirow{4}*{\rotatebox{90}{Electricity}}& 96    & \textcolor{red}{\textbf{0.123}} & \textcolor{red}{\textbf{0.219}} & 0.131 & 0.228 & \textcolor{blue}{0.129} & \textcolor{blue}{0.222} & 0.140 & 0.237 & 0.166 & 0.293 & 0.196 & 0.313 & 0.304 & 0.393 & 0.386 & 0.449 \\
&\multicolumn{1}{c|}{}& 192   & \textcolor{red}{\textbf{0.142}} & \textcolor{red}{\textbf{0.237}} & 0.155 & 0.249 & \textcolor{blue}{0.148} & \textcolor{blue}{0.240} & 0.153 & 0.249 & 0.187 & 0.302 & 0.211 & 0.324 & 0.327 & 0.417 & 0.386 & 0.443  \\
&\multicolumn{1}{c|}{}& 336   & \textcolor{red}{\textbf{0.158}} & \textcolor{red}{\textbf{0.254}} & 0.170 & 0.266 & \textcolor{blue}{0.166} & \textcolor{blue}{0.259} & 0.169 & 0.267 & 0.205 & 0.324 & 0.214 & 0.327 & 0.333 & 0.422 & 0.378 & 0.443 \\
&\multicolumn{1}{c|}{}& 720   & \textcolor{red}{\textbf{0.186}} & \textcolor{red}{\textbf{0.280}} & \textcolor{blue}{0.207} & 0.300 & 0.210 & \textcolor{blue}{0.298} & 0.203 & 0.301 & 0.211 & 0.338 & 0.236 & 0.342 & 0.351 & 0.427 & 0.376 & 0.445 \\

\cline{1-19}
&\multirow{4}*{\rotatebox{90}{Illness}}& 24    & \textcolor{red}{\textbf{1.638}} & \textcolor{blue}{\textbf{0.833}} & 2.060 & 0.960 & \textcolor{blue}{1.816} & \textcolor{red}{0.819} & 2.215 & 1.081 & 2.424 & 1.045 & 2.906 & 1.182 & 4.657 & 1.449  & 1.420 & 2.012 \\
&\multicolumn{1}{c|}{}& 36    & \textcolor{red}{\textbf{1.707}} & \textcolor{red}{\textbf{0.854}} & 2.151 & 1.020 & \textcolor{blue}{2.098} & \textcolor{blue}{0.978} & 1.963 & 0.963 & 2.411 & 1.011 & 2.585 & 1.038 & 4.650 & 1.463 & 7.394 & 2.031 \\
&\multicolumn{1}{c|}{}& 48    & \textcolor{red}{\textbf{1.699}} & \textcolor{red}{\textbf{0.877}} & 2.060 & 0.990 & \textcolor{blue}{1.735} & \textcolor{blue}{0.892} & 2.130 & 1.024 & 2.438 & 1.028 & 3.024 & 1.145 & 5.004 & 1.542 & 7.551 & 2.057  \\
&\multicolumn{1}{c|}{}& 60    & \textcolor{red}{\textbf{1.555}} & \textcolor{red}{\textbf{0.812}} & 2.220 & 1.030 & \textcolor{blue}{1.578} & \textcolor{blue}{0.818} & 2.368 & 1.096 & 2.442 & 1.022 & 2.761 & 1.114 & 5.071 & 1.543 & 7.662 & 2.100 \\
\cline{1-19}
&\multirow{4}*{\rotatebox{90}{ETTh1}}& 96    & \textcolor{red}{\textbf{0.368}} & \textcolor{red}{\textbf{0.395}} & 0.399 & 0.417 & \textcolor{blue}{0.378} & \textcolor{blue}{0.400} & 0.375 & 0.399 & 0.386 & 0.425 & 0.435 & 0.446 & 0.941 & 0.769 & 0.664 & 0.612  \\
&\multicolumn{1}{c|}{}& 192   & \textcolor{red}{\textbf{0.409}} & \textcolor{red}{\textbf{0.418}} & 0.442 & 0.446 & \textcolor{blue}{0.414} & \textcolor{blue}{0.421} & 0.405 & 0.416 & 0.434 & 0.456 & 0.456 & 0.457 & 1.007 & 0.786 & 0.790 & 0.681  \\
&\multicolumn{1}{c|}{}& 336   & \textcolor{red}{\textbf{0.436}} & \textcolor{red}{\textbf{0.440}} & 0.463 & 0.462 & \textcolor{blue}{0.440} & \textcolor{red}{0.440} & 0.439 & \textcolor{blue}{0.443} & 0.481 & 0.472 & 0.486 & 0.487 & 1.038 & 0.784 & 0.891 & 0.738  \\
&\multicolumn{1}{c|}{}& 720   & \textcolor{red}{\textbf{0.451}} & \textcolor{red}{\textbf{0.464}} & 0.496 & 0.501 & \textcolor{blue}{0.456} & \textcolor{blue}{0.471} & 0.472 & 0.490 & 0.481 & 0.512 & 0.515 & 0.517 & 1.144 & 0.857 & 0.963 & 0.782 \\

	\cline{1-19}
&\multirow{4}*{\rotatebox{90}{ETTm1}}& 96    & \textcolor{red}{\textbf{0.279}} & \textcolor{red}{\textbf{0.339}} & 0.302 & 0.356 & \textcolor{blue}{0.292} & \textcolor{blue}{0.342} & 0.299 & 0.343 & 0.316 & 0.380 & 0.510 & 0.492 & 0.626 & 0.560 & 0.543 & 0.510 \\
&\multicolumn{1}{c|}{}& 192   & \textcolor{red}{\textbf{0.327}} & \textcolor{blue}{\textbf{0.368}} & 0.344 & 0.382 & \textcolor{blue}{0.330} & \textcolor{blue}{0.368} & 0.335 & \textcolor{red}{0.365} & 0.361 & 0.409 & 0.514 & 0.495 & 0.725 & 0.619 & 0.557 & 0.537 \\
&\multicolumn{1}{c|}{}& 336   & \textcolor{red}{\textbf{0.360}} & \textcolor{red}{\textbf{0.388}} & 0.378 & 0.403 & \textcolor{blue}{0.365} & \textcolor{blue}{0.391} & 0.369 & 0.386 & 0.392 & 0.425 & 0.510 & 0.492 & 1.005 & 0.741 & 0.754 & 0.655 \\
&\multicolumn{1}{c|}{}& 720   & \textcolor{red}{\textbf{0.417}} & \textcolor{red}{\textbf{0.428}} & 0.438 & 0.437 & \textcolor{red}{0.417} & 0.424 & 0.425 & \textcolor{blue}{0.423} & 0.446 & 0.458 & 0.527 & 0.493 & 1.133 & 0.845 & 0.908 & 0.724 \\

		\cline{1-19}
&\multirow{4}*{\rotatebox{90}{Solar}}& 96    & \textcolor{red}{\textbf{0.169}} & \textcolor{red}{\textbf{0.232}} & 0.196 & \textcolor{blue}{0.249} & \textcolor{blue}{0.194 }& \textcolor{blue}{0.249} & 0.221 & 0.289 & 0.241 & 0.299 & 0.266 & 0.311 & 0.208 & 0.237 & 0.228 & 0.249 \\
&\multicolumn{1}{c|}{}& 192   & \textcolor{red}{\textbf{0.184}} & \textcolor{red}{\textbf{0.241}} & \textcolor{blue}{0.219} & 0.266 & 0.228 & \textcolor{blue}{0.261} & 0.249 & 0.285 & 0.268 & 0.314 & 0.271 & 0.315 & 0.229 & 0.259 & 0.237 & 0.268 \\
&\multicolumn{1}{c|}{}& 336   & \textcolor{red}{\textbf{0.193}} & \textcolor{red}{\textbf{0.250}} & \textcolor{blue}{0.218} & 0.266 & 0.221 &\textcolor{blue}{ 0.254} & 0.263 & 0.291 & 0.288 & 0.311 & 0.281 & 0.317 & 0.235 & 0.272 & 0.247 & 0.278 \\
&\multicolumn{1}{c|}{}& 720   & \textcolor{red}{\textbf{0.203}} & \textcolor{red}{\textbf{0.261}} & 0.230 & 0.279 & \textcolor{blue}{0.219} & \textcolor{blue}{0.271} & 0.244 & 0.296 & 0.271 & 0.315 & 0.295 & 0.319 & 0.233 & 0.275 & 0.255 & 0.286 \\

			\bottomrule
		\end{tabular}
	}
	\label{tab:supervised}
\end{table*}

\textbf{Main results.} 
The comprehensive forecasting results are detailed in Table~\ref{tab:supervised}, with the top-performing models highlighted in red and the second-best in blue. 
A lower MSE/MAE signifies more accurate predictions, and in this context, the proposed GridTST consistently delivers state-of-the-art performance. 
Notably, PatchTST, previously the leading model on the Electricity and Weather datasets, struggles in several Traffic dataset scenarios. 
This is likely due to the dataset's large number of variates, where PatchTST's patching mechanism might lose focus on specific localities, thus failing to handle rapid fluctuations effectively.
In contrast, the iTransformer stands out in the traffic datasets, adeptly handling up to 800 variates. 
However, it falls short in other datasets where the number of variates is fewer and time information plays a more crucial role. 
This disparity highlights the iTransformer's limitations in contexts where temporal dynamics are more significant.

Our proposed method enhances this by modeling both temporal and variate aspects through bidirectional attentions. 
For example, it exhibits superior performance in 26 out of 28 tasks when evaluated using two metrics.
By aggregating the entire series of time and variate variations for representation, it can more adeptly manage such forecasting challenges. 
Consequently, native Transformer components prove effective for temporal modeling and multivariate correlation, and our proposed bidirectional architecture is well-suited for tackling complex real-world time series forecasting scenarios.

\begin{figure*}[ht]
\centering
\includegraphics[width=\linewidth]{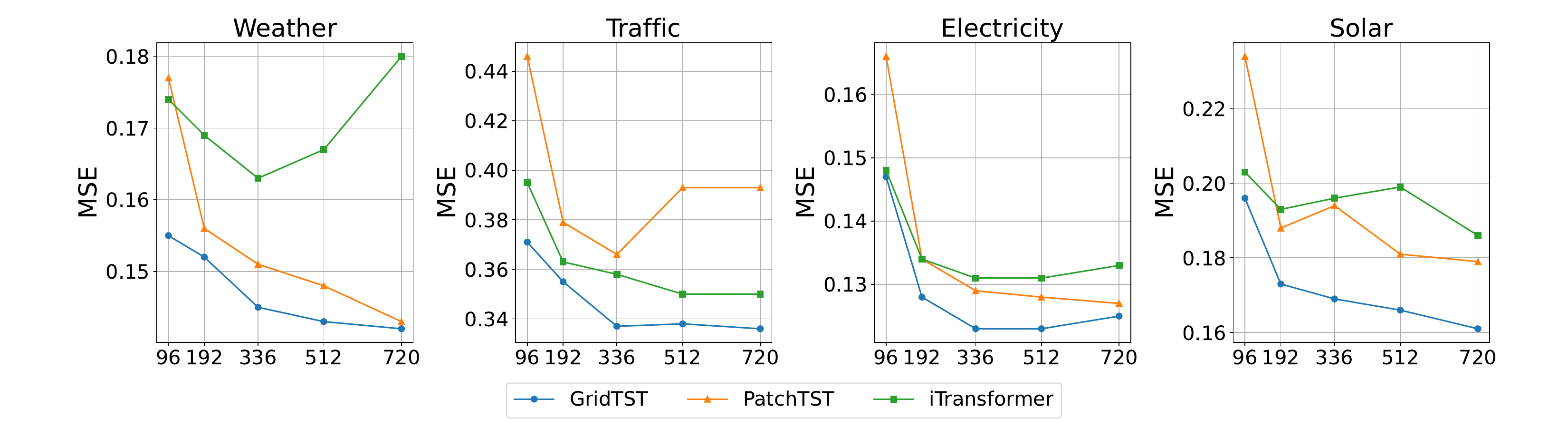}
\caption{ Forecasting Performance with Lookback Length \( T \in \{96, 172, 336, 512, 720\} \) and Fixed Prediction Length \( F = 96 \).
The performance of time-centric PatchTST or variate-centric iTransformer forecasters does not markedly improve with increased lookback length. 
In contrast, our GridTST framework enhances the vanilla Transformer, yielding improved performance when utilizing an enlarged lookback window.}
\label{fig:lookback}
\end{figure*}

\textbf{Increasing lookback length.}
Numerous prior studies~\cite{nie2022time} have noted that increasing the lookback length does not invariably enhance forecasting performance in Transformers, a phenomenon often attributed to attention becoming dispersed over an expanding input sequence. 
In this context, we assessed the performance of our GridTST, along with time-centric PatchTST and variate-centric iTransformer, as illustrated in Figure~\ref{fig:lookback}, particularly focusing on scenarios with extended lookback lengths.
The full data can be found in Table~\ref{tab:look96} to Table~\ref{tab:look720} in Appendix.
It is apparent that the variate-centric iTransformer exhibits subpar performance on the Weather dataset, which comprises a limited number of variates, while the time-centric PatchTST underperforms on the Traffic dataset, characterized by hundreds of variates. 
Conversely, our model demonstrates consistent enhancement in performance with the expansion of the lookback window size, irrespective of whether the number of variates is large or small.
This improvement is particularly notable in the traffic and electricity datasets, where the number of variate channels is substantial. 
These findings validate the effectiveness of applying bidirectional attention across both temporal and variate dimensions, enabling Transformers to harness an extended lookback window. 
This approach effectively models complex relationships among hundreds of variables, thereby yielding more precise predictions.
It's crucial to note that PatchTST and GridTST both utilize patching, whereas iTransformer relies on mapping. 
Our patch-based methods provide flexibility by allowing us to adjust patch lengths based on the lookback window size. 
This adaptability is a distinct advantage compared to mapping-based methods, which lack the capability to tailor their approach for different lookback windows.

\begin{figure*}[ht]
\centering
\includegraphics[width=\linewidth]{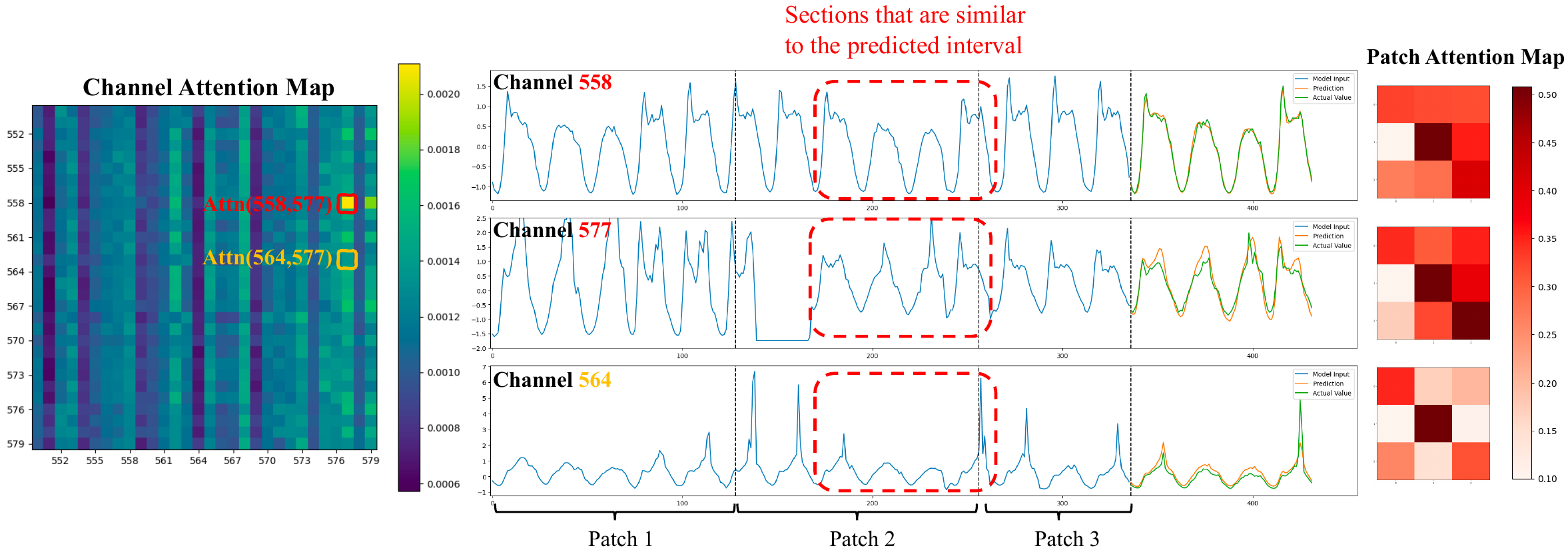}
\caption{ Visualization of attention maps and time series forecasts from the Traffic dataset. For each time series, the input data is represented in blue, the GridTST model's predictions in orange, and the actual observed values in green. The three black demarcation lines indicate the three segmented patches. The synergy of horizontal and vertical attention mechanisms enables the model to refine its forecasts by concentrating on the spatial and temporal information deemed most pertinent.}
\label{fig:analysis}
\end{figure*}

\subsection{Visualization Case Study}
As shown in Figure~\ref{fig:analysis}, we present an intuitive visualization of the multivariate and multi-time-step correlations via attention maps and the forecasting outcomes for three selected time series from the Traffic dataset, processed using GridTST in the channel-first arrangement. The model configuration includes a lookback length of 336, a fixed prediction length of 96, and a patch length also set to 96. The attention maps depicted are the result of averaging the attention matrices across all heads.

The channel attention map on the left highlights discernible correlations that mirror the inherent trends in the raw lookback series. Notably, time series 558 and 577 exhibit more closely aligned trends in contrast to series 564 and 577, which are not highlighted.
This observation substantiates the model's proficiency in detecting inter-channel similarities by leveraging vertical attention mechanisms. Such mechanisms enable mutual reinforcement between analogous time series, thereby enhancing the prediction capabilities by leveraging the diversity of the channels.

Delving into the patch attention maps on the right, we observe that time series with similar trends share similar patch attention distributions. Furthermore, these maps frequently emphasize segments that mirror the predicted intervals, as delineated by the red dashed lines. This tendency underscores the significance of historical patterns within the data, and the horizontal attention's adeptness at capturing these temporal relationships.

The central temporal graphs demonstrate the model's adeptness at pinpointing critical time patches while simultaneously conducting inter-channel comparisons. The synergy of horizontal and vertical attention mechanisms enables the model to forecast by focusing on the spatial and temporal information deemed most pertinent.


\subsection{Ablation study}
\label{ablation}

\begin{table*}
  \centering
  \caption{Ablation study of GridTST}
  \scriptsize
  \begin{tabular}{cc|c|cc|cc|cc}
      \cline{2-9}
      & \multicolumn{2}{c|}{Models} & \multicolumn{2}{c|}{Channel First} & \multicolumn{2}{c|}{Time First} & \multicolumn{2}{c}{Alternate} \\
      \cline{2-9}
      & \multicolumn{2}{c|}{Metric} & MSE & MAE & MSE & MAE & MSE & MAE \\
      \cline{2-9}
         & \multirow{4}*{\rotatebox{90}{Weather}} & 96    & \textbf{0.145} & \textbf{0.195} & 0.149 & 0.198 & 0.146 & 0.196 \\
  & \multicolumn{1}{c|}{} & 192   & 0.191 & 0.240  & 0.193 & \textbf{0.240}  & \textbf{0.190}  & \textbf{0.238} \\
  & \multicolumn{1}{c|}{} & 336   & 0.243 & 0.280  & 0.241 & \textbf{0.277} & \textbf{0.240}  & 0.278 \\
  & \multicolumn{1}{c|}{} & 720   & 0.316 & 0.333 & \textbf{0.315} & \textbf{0.331} & 0.335 & 0.342 \\
  \cline{2-9} 
  & \multirow{4}*{\rotatebox{90}{Traffic}} & 96    & \textbf{0.337} & \textbf{0.241} & 0.352 & 0.254 & 0.351 & 0.250 \\
  & \multicolumn{1}{c|}{} & 192   & \textbf{0.373} & \textbf{0.259} & 0.382 & 0.270 & 0.381 & 0.267 \\
  & \multicolumn{1}{c|}{} & 336   & \textbf{0.378} & \textbf{0.260} & 0.386 & 0.272 & 0.384 & 0.267 \\
  & \multicolumn{1}{c|}{} & 720   & \textbf{0.403} & \textbf{0.276} & 0.414 & 0.285 & 0.414 & 0.285 \\
  \cline{2-9}
  & \multirow{4}*{\rotatebox{90}{Electricity}} & 96    & \textbf{0.123} & \textbf{0.219} & 0.125 & 0.222 & 0.125 & 0.222 \\
  & \multicolumn{1}{c|}{} & 192   & \textbf{0.142} & \textbf{0.237} & 0.146 & 0.242 & 0.145 & 0.241 \\
  & \multicolumn{1}{c|}{} & 336   & \textbf{0.158} & \textbf{0.254}& 0.166 & 0.263 & 0.158 & 0.255 \\
  & \multicolumn{1}{c|}{} & 720   & 0.206 & 0.295 & 0.191 & 0.286 & \textbf{0.186} & \textbf{0.280} \\
  \cline{2-9}
  & \multirow{4}*{\rotatebox{90}{ETTh1}} & 96    & \textbf{0.368} & \textbf{0.395} & 0.374 & 0.397 & 0.370 & 0.396 \\
  & \multicolumn{1}{c|}{} & 192   & \textbf{0.409} & \textbf{0.418} & 0.413 & 0.419 & 0.412 & 0.420 \\
  & \multicolumn{1}{c|}{} & 336   & \textbf{0.436} & 0.440  & \textbf{0.436} & \textbf{0.439} & 0.439 & 0.440 \\
  & \multicolumn{1}{c|}{} & 720   & 0.462 & 0.474 & \textbf{0.450}  & \textbf{0.464} & 0.451 & \textbf{0.464} \\
           \cline{2-9}
  \end{tabular}
  \label{tab:ablation}
\end{table*}
Table~\ref{tab:ablation} presents an ablation study on various  attention sequence arrangements, offering valuable insights into the performance dynamics of different models. 
It is evident that all models achieve relatively good performance, underscoring the robustness and consistency of our proposed model. 
Specifically, the channel-first arrangement emerges as the most effective, outperforming other configurations in more metrics and settings. 
This approach aligns intuitively with human cognitive processes, where we typically analyze variables at a single time point before extending our analysis along the timeline. 
We select attention strategy for different dataset by the their performance on the validation set.

\subsection{Scalability of GridTST}
\begin{table}[ht]
\centering
\caption{Speedup ratio of our framework with BetterTransformer Library.}
\label{better}
\resizebox{0.6\columnwidth}{!}{%
\begin{tabular}{@{}lccccccc@{}}
\toprule
Input\_length & 64   & 128  & 256  & 512  & 1024 & 2048 & 4096 \\ 
\midrule
GPU setting          & x1.49 & x1.62 & x1.62 & x1.80 & x2.04 & x2.46 & x2.62 \\
CPU setting & x1.19 & x1.54 & x1.34 & x1.21 & x1.17 & x1.24 & x1.20 \\
\bottomrule
\end{tabular}%
}
\end{table}

One advantage of adopting the vanilla Transformer architecture is the ability to tap into the established ecosystem that has been purposefully developed to optimize Transformer performance on CPUs and GPUs. 
By leveraging specialized techniques such as sparsity and fused kernels, significant speedups can be achieved. 
For instance, using the BetterTransformer library, a single line of code can dramatically accelerate processing times of our GridTST by utilizing nested tensors for sparsity and integrating flash-attention \cite{dao2022flashattention}. 
This convenience is not available for various heavily modified Transformer structures \cite{zhang2022crossformer,liu2021pyraformer}, and thus cannot benefit from the existing ecosystem. 
As demonstrated in Table~\ref{better}, on GPUs, the speedup increases with the length of the input sequence, indicating that GridTST is particularly effective when scaling to extremely long time series.
On CPUs, the acceleration remains consistent, making it a viable option for deployment scenarios.

\subsection{Efficient training strategy}
\label{section:efficient_training_strategy}

  \begin{figure*}[ht]
\centering
\includegraphics[width=1\linewidth]{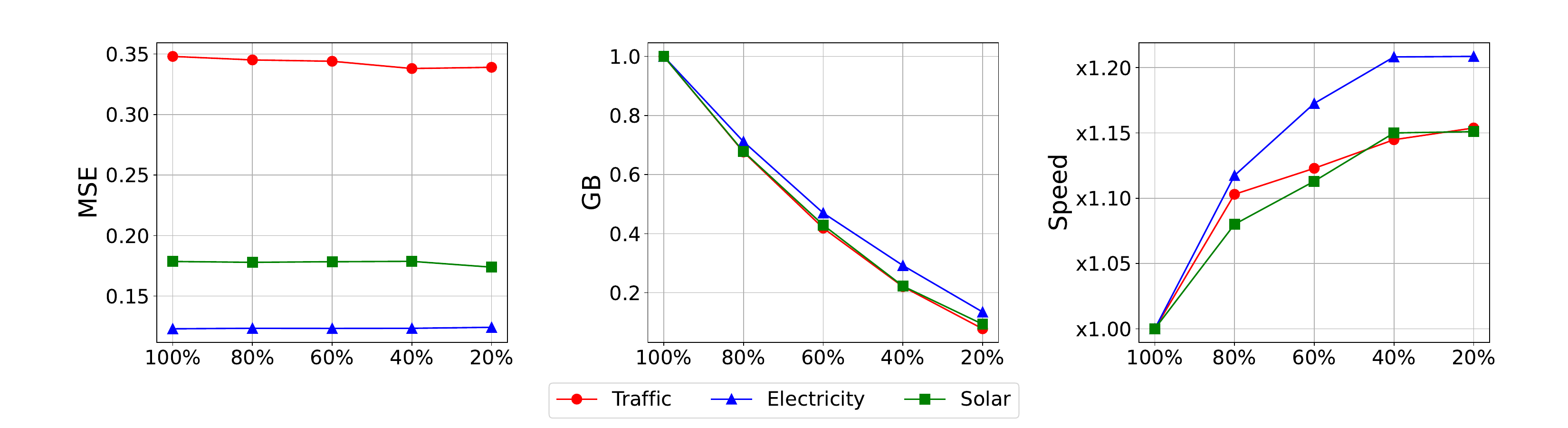}
\caption{ Analysis of the Proposed Training Strategy.
The performance (Left) maintains stable across partially trained variants of each batch, with the sampled ratio ranging from 20\% to 100\%. 
Concurrently, there is a notable reduction in both the memory footprint (Middle) and the latency (Right) of the training process.
}
\label{fig:efficiency}
\end{figure*}

In the Transformer architecture, the self-attention mechanism's quadratic complexity can become burdensome during training as the number of variables increases. 
One approach to mitigate this complexity is to randomly select a subset of variables for each training batch, focusing the model's training on these chosen variables. 
In Figure~\ref{fig:efficiency}, we demonstrate the effectiveness of this method by training the model with a randomly sampled set of variables and evaluating its performance. 
Full data can be found in Table~\ref{tab:appendix_ratio} in Appendix.
The results reveal that our proposed strategy achieves performance comparable to training with the full set of variables, while significantly reducing memory usage and increasing training speed. 
This underscores the effectiveness of our vertical attention approach in capturing variable dependencies from a limited set of training samples.

 \section{Conclusion and Broader Impacts}
	\label{conclusion}
In our study, we introduced GridTST, an innovative model for time series prediction with broad applications in fields like finance, economics, climate, and healthcare.
GridTST combines the strengths of two prevailing approaches by treating time series data as a grid, incorporating both time and variate dimensions.
Our model employs horizontal and vertical attentions to efficiently capture temporal and multivariate correlations, enhancing analytical capabilities. 
We consistently achieved state-of-the-art performance on real-world datasets, showcasing GridTST's effectiveness.
For future work, we aim to delve into another direction by extending GridTST's capabilities to address multi-modal time series data.
\bibliographystyle{unsrt}
\bibliography{sample-base}

\appendix

\clearpage
\section{Dataset}
\label{appendix_data}
\begin{table*}[ht]
\centering
\caption{Detailed dataset descriptions. Channels denotes the variate number of each dataset. 
Dataset Partition denotes the ratio number of time points in (Train, Validation, Test) split respectively. 
Prediction Length denotes the future time points to be predict and four prediction settings are included in each dataset. Frequency denotes the sampling interval of time points.}
\resizebox{\textwidth}{!}{\begin{tabular}{lcccccc}
\toprule
\textbf{Dataset} & \textbf{\# Channels} & \textbf{\# TimeSteps} & \textbf{Prediction Length} & \textbf{Dataset Partition} & \textbf{Frequency} & \textbf{Category} \\
\hline
Weather    & 21  & 52696 & \{96,192,336,720\} & 7:1:2 & 10min & Weather \\
Traffic    & 862 & 17544 & \{96,192,336,720\} & 7:1:2 & Hourly & Transportation \\
Electricity & 321 & 26304 & \{96,192,336,720\} & 7:1:2 & Hourly & Electricity \\
Illness    & 7   & 966   & \{12,24,48,60\}    & 7:1:2 & Weekly & Illness \\
Etth1      & 7   & 17420 & \{96,192,336,720\} & 6:2:2 & Hourly & Electricity \\
Ettm1      & 7   & 69680 & \{96,192,336,720\} & 6:2:2 & 15min & Electricity \\
Solar      & 137 & 52560 & \{96,192,336,720\} & 7:1:2 & 10min & Energy \\
\bottomrule
\end{tabular}}
\label{tab:datasets}
\end{table*}

We conduct experiments on seven real-world datasets to evaluate the performance of the proposed GridTST. The first four datasets are collected by \cite{wu2021autoformer}: \textit{Weather} includes 21 meteorological factors collected every 10 minutes from the Weather Station of the Max Planck Biogeochemistry Institute in 2020; \textit{Traffic} collects hourly road occupancy rates measured by 862 sensors of the San Francisco Bay area freeways from January 2015 to December 2016; \textit{Electricity} records the hourly electricity consumption data of 321 clients; \textit{Illness} dataset collects the number of patients and influenza-like illness ratio on a weekly frequency. 
 Additionally, \textit{ETTh1} and \textit{Ettm1} by \cite{chen2023tsmixer} contain 7 factors of electricity transformer from July 2016 to July 2018, recorded hourly. 
 Finally, \textit{Solar-Energy} by \cite{lai2018modeling} records the solar power production of 137 PV plants in 2006, sampled every 10 minutes.
 
\section{Comparison with existing baselines}

\begin{table}[h!]
\centering

\caption{Comparison of Transformer Models}
\label{compare}
\begin{tabular}{l|ccc}
\hline
& \textbf{Vanilla} & \textbf{Multivariate} & \textbf{Sequential} \\
& \textbf{Transformer} & \textbf{Modeling} & \textbf{Modeling} \\ \hline
\textbf{DLinear }        & \textcolor{red}{\texttimes} & \textcolor{red}{\texttimes} & \textcolor{red}{\texttimes} \\ 
\textbf{CrossFormer }    & \textcolor{red}{\texttimes} & \checkmark                   & \checkmark                   \\ 
\textbf{PatchTST }       & \checkmark                  & \textcolor{red}{\texttimes} & \checkmark                   \\ 
\textbf{iTransformer}  & \checkmark                  & \checkmark                  & \textcolor{red}{\texttimes} \\
\textbf{GridTST}                   & \checkmark                  & \checkmark                  & \checkmark                   \\ \hline
\end{tabular}
\end{table}

Table~\ref{compare} presents a clear comparison of various transformer models, with GridTST emerging as the most versatile. Unlike DLinear, CrossFormer, and PatchTST, GridTST supports both multivariate and sequential modeling, alongside leveraging the foundational vanilla Transformer architecture. This dual capability allows GridTST to effectively capture the intricate dynamics of time series data. While iTransformer also utilizes the vanilla Transformer framework and multivariate modeling, it falls short in sequential modeling, which GridTST accommodates. Therefore, GridTST stands out for its comprehensive approach, making it well-suited for complex time series analysis that demands capturing both simultaneous variable relationships and sequential dependencies.

\begin{table*}[ht!]\centering
\begin{minipage}{0.99\columnwidth}\vspace{0mm}    \centering
\begin{tcolorbox} 
    \centering
     \hspace{-6mm}
\begin{lstlisting}[language=Python, caption={GridTST forward pass with alternate attention.\label{code}}, basicstyle=\footnotesize\ttfamily, breaklines=true, linewidth=\textwidth, xleftmargin=0pt]
def gridtst_forward_with_alternate_attention(time_series, GridTSTLayers):
    """
    The input time_series is a 4D tensor of shape: [batch_size, num_variates, num_patches,
    d_model]
    """
    batch_size, num_variates, num_patches, d_model = time_series.shape
    time_series = time_series.view(batch_size * num_variates, num_patches, d_model)

    for idx, layer in enumerate(GridTSTLayers):
        if idx % 2 == 0:
            time_series = layer(time_series)
        else:
            time_series = time_series.reshape(-1, num_variates, num_patches, d_model)
            time_series = time_series.permute(0, 2, 1, 3)
            time_series = time_series.reshape(-1, num_variates, d_model)

            time_series = layer(time_series)

            time_series = time_series.reshape(-1, num_patches, num_variates, d_model)
            time_series = time_series.permute(0, 2, 1, 3)
            time_series = time_series.reshape(-1, num_patches, d_model)
    return time_series
\end{lstlisting}
\end{tcolorbox}
\vspace{-2mm}
\label{table:paraphrase_instructions}
\end{minipage}
\end{table*}

\section{Increasing lookback length}

The performance of data with different lookback lengths is available in Table~\ref{tab:look96} to Table~\ref{tab:look720}.


\begin{table}[H]
\centering
\caption{Performance of GridTST, iTransformer and PatchTST when using lookback windows of different lengths, The prediction length is fixed at 96. The performance is measured by MSE.}
\begin{tabular}{cc|c|cc|cc|cc}
      \cline{2-9}
      & \multicolumn{2}{c|}{Models} & \multicolumn{2}{c|}{Channel First} & \multicolumn{2}{c|}{Time First} & \multicolumn{2}{c}{Alternate} \\
      \cline{2-9}
      & \multicolumn{2}{c|}{Metric} & MSE & MAE & MSE & MAE & MSE & MAE \\
      \cline{2-9}
         & \multirow{4}*{\rotatebox{90}{Weather}} & 96    & 0.155 & 0.2 & 0.177 & 0.218 & 0.174 & 0.214 \\
  & \multicolumn{1}{c|}{} & 192   & 0.206 & 0.247 & 0.225 & 0.259 & 0.221 & 0.254 \\
  & \multicolumn{1}{c|}{} & 336   & 0.264 & 0.289 & 0.278 & 0.297 & 0.278 & 0.296 \\
  & \multicolumn{1}{c|}{} & 720   & 0.343 & 0.343 & 0.354 & 0.348 & 0.358 & 0.349 \\
  \cline{2-9} 
  & \multirow{4}*{\rotatebox{90}{Traffic}} & 96    & 0.371 & 0.247 & 0.446 & 0.283 & 0.395 & 0.268 \\
  & \multicolumn{1}{c|}{} & 192   & 0.396 & 0.256 & 0.452 & 0.285 & 0.417 & 0.276 \\
  & \multicolumn{1}{c|}{} & 336   & 0.406 & 0.261 & 0.467 & 0.291 & 0.433 & 0.283 \\
  & \multicolumn{1}{c|}{} & 720   & 0.436 & 0.282 & 0.5 & 0.309 & 0.467 & 0.302 \\
  \cline{2-9}
  & \multirow{4}*{\rotatebox{90}{Electricity}} & 96    & 0.147 & 0.239 & 0.166 & 0.252 & 0.148 & 0.24 \\
  & \multicolumn{1}{c|}{} & 192   & 0.162 & 0.256 & 0.175 & 0.261 & 0.162 & 0.253 \\
  & \multicolumn{1}{c|}{} & 336   & 0.178 & 0.272 & 0.19 & 0.277 & 0.178 & 0.269 \\
  & \multicolumn{1}{c|}{} & 720   & 0.214 & 0.305 & 0.23 & 0.311 & 0.225 & 0.317 \\
  \cline{2-9}
  & \multirow{4}*{\rotatebox{90}{Solar}} & 96    & 0.196 & 0.242 & 0.234 & 0.286 & 0.203 & 0.235 \\
  & \multicolumn{1}{c|}{} & 192   & 0.225 & 0.265 & 0.267 & 0.31 & 0.233 & 0.261 \\
  & \multicolumn{1}{c|}{} & 336   & 0.245 & 0.281 & 0.29 & 0.315 & 0.248 & 0.273 \\
  & \multicolumn{1}{c|}{} & 720   & 0.251 & 0.291 & 0.289 & 0.317 & 0.249 & 0.275 \\
           \cline{2-9}
  \end{tabular}
\label{tab:look96}
\end{table}

\begin{table}[H]
\centering
\caption{Performance of GridTST, iTransformer and PatchTST when using lookback windows of different lengths, The prediction length is fixed at 192. The performance is measured by MSE.}
\begin{tabular}{cc|c|cc|cc|cc}
      \cline{2-9}
      & \multicolumn{2}{c|}{Models} & \multicolumn{2}{c|}{Channel First} & \multicolumn{2}{c|}{Time First} & \multicolumn{2}{c}{Alternate} \\
      \cline{2-9}
      & \multicolumn{2}{c|}{Metric} & MSE & MAE & MSE & MAE & MSE & MAE \\
      \cline{2-9}
         & \multirow{4}*{\rotatebox{90}{Weather}} & 96    & 0.152 & 0.199 & 0.156 & 0.201 & 0.169 & 0.216 \\
  & \multicolumn{1}{c|}{} & 192   & 0.195 & 0.241 & 0.202 & 0.243 & 0.214 & 0.255 \\
  & \multicolumn{1}{c|}{} & 336   & 0.252 & 0.285 & 0.252 & 0.285 & 0.266 & 0.293 \\
  & \multicolumn{1}{c|}{} & 720   & 0.329 & 0.339 & 0.331 & 0.336 & 0.341 & 0.345 \\
  \cline{2-9} 
  & \multirow{4}*{\rotatebox{90}{Traffic}} & 96    & 0.355 & 0.249 & 0.379 & 0.253 & 0.363 & 0.258 \\
  & \multicolumn{1}{c|}{} & 192   & 0.385 & 0.264 & 0.398 & 0.261 & 0.39 & 0.268 \\
  & \multicolumn{1}{c|}{} & 336   & 0.397 & 0.274 & 0.411 & 0.268 & 0.399 & 0.282 \\
  & \multicolumn{1}{c|}{} & 720   & 0.424 & 0.291 & 0.443 & 0.288 & 0.434 & 0.294 \\
  \cline{2-9}
  & \multirow{4}*{\rotatebox{90}{Electricity}} & 96    & 0.128 & 0.223 & 0.134 & 0.225 & 0.134 & 0.228 \\
  & \multicolumn{1}{c|}{} & 192   & 0.147 & 0.241 & 0.151 & 0.242 & 0.156 & 0.25 \\
  & \multicolumn{1}{c|}{} & 336   & 0.16 & 0.256 & 0.169 & 0.262 & 0.171 & 0.266 \\
  & \multicolumn{1}{c|}{} & 720   & 0.19 & 0.283 & 0.206 & 0.295 & 0.195 & 0.288 \\
  \cline{2-9}
  & \multirow{4}*{\rotatebox{90}{Solar}} & 96    & 0.173 & 0.23 & 0.188 & 0.241 & 0.193 & 0.232 \\
  & \multicolumn{1}{c|}{} & 192   & 0.196 & 0.262 & 0.221 & 0.256 & 0.223 & 0.26 \\
  & \multicolumn{1}{c|}{} & 336   & 0.211 & 0.263 & 0.228 & 0.264 & 0.231 & 0.271 \\
  & \multicolumn{1}{c|}{} & 720   & 0.214 & 0.266 & 0.231 & 0.274 & 0.227 & 0.27 \\
           \cline{2-9}
  \end{tabular}
\label{tab:look192}
\end{table}

\begin{table}[H]
\centering
\caption{Performance of GridTST, iTransformer and PatchTST when using lookback windows of different lengths, The prediction length is fixed at 512. The performance is measured by MSE.}
\begin{tabular}{cc|c|cc|cc|cc}
      \cline{2-9}
      & \multicolumn{2}{c|}{Models} & \multicolumn{2}{c|}{Channel First} & \multicolumn{2}{c|}{Time First} & \multicolumn{2}{c}{Alternate} \\
      \cline{2-9}
      & \multicolumn{2}{c|}{Metric} & MSE & MAE & MSE & MAE & MSE & MAE \\
      \cline{2-9}
         & \multirow{4}*{\rotatebox{90}{Weather}} & 96    & 0.143 & 0.194 & 0.148 & 0.198 & 0.167 & 0.219 \\
  & \multicolumn{1}{c|}{} & 192   & 0.189 & 0.238 & 0.192 & 0.24 & 0.208 & 0.254 \\
  & \multicolumn{1}{c|}{} & 336   & 0.241 & 0.278 & 0.243 & 0.28 & 0.266 & 0.294 \\
  & \multicolumn{1}{c|}{} & 720   & 0.313 & 0.33 & 0.313 & 0.332 & 0.341 & 0.345 \\
  \cline{2-9} 
  & \multirow{4}*{\rotatebox{90}{Traffic}} & 96    & 0.338 & 0.24 & 0.393 & 0.283 & 0.35 & 0.256 \\
  & \multicolumn{1}{c|}{} & 192   & 0.358 & 0.251 & 0.408 & 0.29 & 0.376 & 0.268 \\
  & \multicolumn{1}{c|}{} & 336   & 0.376 & 0.261 & 0.419 & 0.294 & 0.386 & 0.274 \\
  & \multicolumn{1}{c|}{} & 720   & 0.399 & 0.274 & 0.452 & 0.313 & 0.417 & 0.289 \\
  \cline{2-9}
  & \multirow{4}*{\rotatebox{90}{Electricity}} & 96    & 0.123 & 0.22 & 0.128 & 0.221 & 0.131 & 0.227 \\
  & \multicolumn{1}{c|}{} & 192   & 0.144 & 0.24 & 0.149 & 0.244 & 0.153 & 0.25 \\
  & \multicolumn{1}{c|}{} & 336   & 0.162 & 0.258 & 0.163 & 0.259 & 0.168 & 0.264 \\
  & \multicolumn{1}{c|}{} & 720   & 0.194 & 0.285 & 0.199 & 0.291 & 0.191 & 0.284 \\
  \cline{2-9}
  & \multirow{4}*{\rotatebox{90}{Solar}} & 96    & 0.166 & 0.232 & 0.181 & 0.239 & 0.199 & 0.242 \\
  & \multicolumn{1}{c|}{} & 192   & 0.174 & 0.242 & 0.187 & 0.256 & 0.207 & 0.269 \\
  & \multicolumn{1}{c|}{} & 336   & 0.176 & 0.244 & 0.202 & 0.267 & 0.223 & 0.266 \\
  & \multicolumn{1}{c|}{} & 720   & 0.212 & 0.275 & 0.221 & 0.287 & 0.223 & 0.284 \\
           \cline{2-9}
  \end{tabular}
\label{tab:look512}
\end{table}

\begin{table}[H]
\centering
\caption{Performance of GridTST, iTransformer and PatchTST when using lookback windows of different lengths, The prediction length is fixed at 720. The performance is measured by MSE.}
\begin{tabular}{cc|c|cc|cc|cc}
      \cline{2-9}
      & \multicolumn{2}{c|}{Models} & \multicolumn{2}{c|}{Channel First} & \multicolumn{2}{c|}{Time First} & \multicolumn{2}{c}{Alternate} \\
      \cline{2-9}
      & \multicolumn{2}{c|}{Metric} & MSE & MAE & MSE & MAE & MSE & MAE \\
      \cline{2-9}
         & \multirow{4}*{\rotatebox{90}{Weather}} & 96    & 0.142 & 0.194 & 0.143 & 0.193 & 0.18 & 0.231 \\
  & \multicolumn{1}{c|}{} & 192   & 0.188 & 0.238 & 0.188 & 0.239 & 0.225 & 0.266 \\
  & \multicolumn{1}{c|}{} & 336   & 0.239 & 0.277 & 0.242 & 0.283 & 0.285 & 0.311 \\
  & \multicolumn{1}{c|}{} & 720   & 0.314 & 0.331 & 0.306 & 0.327 & 0.352 & 0.357 \\
  \cline{2-9} 
  & \multirow{4}*{\rotatebox{90}{Traffic}} & 96    & 0.336 & 0.244 & 0.393 & 0.288 & 0.35 & 0.256 \\
  & \multicolumn{1}{c|}{} & 192   & 0.351 & 0.251 & 0.402 & 0.29 & 0.364 & 0.265 \\
  & \multicolumn{1}{c|}{} & 336   & 0.376 & 0.263 & 0.418 & 0.297 & 0.383 & 0.272 \\
  & \multicolumn{1}{c|}{} & 720   & 0.438 & 0.304 & 0.456 & 0.317 & 0.41 & 0.286 \\
  \cline{2-9}
  & \multirow{4}*{\rotatebox{90}{Electricity}} & 96    & 0.125 & 0.221 & 0.13 & 0.225 & 0.133 & 0.229 \\
  & \multicolumn{1}{c|}{} & 192   & 0.152 & 0.248 & 0.155 & 0.256 & 0.154 & 0.249 \\
  & \multicolumn{1}{c|}{} & 336   & 0.17 & 0.267 & 0.188 & 0.269 & 0.168 & 0.265 \\
  & \multicolumn{1}{c|}{} & 720   & 0.199 & 0.293 & 0.201 & 0.296 & 0.191 & 0.285 \\
  \cline{2-9}
  & \multirow{4}*{\rotatebox{90}{Solar}} & 96    & 0.161 & 0.23 & 0.171 & 0.232 & 0.186 & 0.237 \\
  & \multicolumn{1}{c|}{} & 192   & 0.189 & 0.267 & 0.199 & 0.269 & 0.202 & 0.27 \\
  & \multicolumn{1}{c|}{} & 336   & 0.202 & 0.276 & 0.211 & 0.279 & 0.217 & 0.283 \\
  & \multicolumn{1}{c|}{} & 720   & 0.229 & 0.306 & 0.231 & 0.298 & 0.221 & 0.283 \\
           \cline{2-9}
  \end{tabular}
\label{tab:look720}
\end{table}

\section{Efficiency}

We show the MAE, MAE, peak memory, and training speed on different datasets with different variant sample ratios in Table~\ref{tab:appendix_ratio}.

\begin{table*}[htb]
\centering
\caption{Ablation study on different variate sample ratios. The lookback window size is 336 and the prediction length is 96.}
\resizebox{\textwidth}{!}{%
\begin{tabular}{c|c|c|c|c|c|c|c|c|c|cc}
\toprule
Variant sample ratio & 0.1 & 0.2 & 0.3 & 0.4 & 0.5 & 0.6 & 0.7 & 0.8 & 0.9 & 1 \\ \hline
\multicolumn{11}{c}{Traffic} \\ \hline
MSE & 0.339 & 0.339 & 0.338 & 0.338 & 0.341 & 0.344 & 0.3467 & 0.3451 & 0.3482 & 0.348 \\
MAE & 0.242 & 0.243 & 0.242 & 0.241 & 0.242 & 0.243 & 0.2464 & 0.2438 & 0.2471 & 0.246 \\
Peak Memory & 0.033 & 0.0782 & 0.1397 & 0.3084 & 0.4189 & 0.5355 & 0.6764 & 0.8251 & 0.9821 & 1 \\
Training Speed & 0.8424 & 0.8667 & 0.9322 & 0.956 & 0.9462 & 0.9732 & 0.9646 & 0.9922 & 0.9813 & 1 \\ \hline
\multicolumn{11}{c}{Electricity} \\ \hline
MSE & 0.1254 & 0.1242 & 0.1237 & 0.1234 & 0.1233 & 0.1234 & 0.1234 & 0.1234 & 0.123 & 0.123 \\
MAE & 0.2197 & 0.2187 & 0.2184 & 0.2186 & 0.2188 & 0.2188 & 0.2191 & 0.219 & 0.219 & 0.219 \\
Peak Memory & 0.0575 & 0.1349 & 0.2077 & 0.2923 & 0.3765 & 0.47 & 0.5986 & 0.7115 & 0.8342 & 1 \\
Training Speed & 0.8492 & 0.8274 & 0.8587 & 0.9245 & 0.9697 & 0.9997 & 0.1003 & 0.9702 & 0.9893 & 1 \\ \hline
\multicolumn{11}{c}{Solar} \\ \hline
MSE & 0.1775 & 0.1739 & 0.18 & 0.1787 & 0.1784 & 0.1777 & 0.1779 & 0.1785 & 0.1786 & 0.1786 \\
MAE & 0.2395 & 0.2368 & 0.2488 & 0.2494 & 0.25 & 0.2495 & 0.2503 & 0.2521 & 0.2511 & 0.2516 \\
Peak Memory & 0.0376 & 0.0936 & 0.152 & 0.2266 & 0.3248 & 0.4293 & 0.5454 & 0.6779 & 0.8317 & 1 \\
Training Speed & 0.8439 & 0.8688 & 0.9263 & 0.9388 & 0.9633 & 0.9671 & 0.9897 & 0.9995 & 1.0078 & 1 \\ \bottomrule
\end{tabular}}
\label{tab:appendix_ratio}
\end{table*}

\section*{NeurIPS Paper Checklist}

\begin{enumerate}

\item {\bf Claims}
    \item[] Question: Do the main claims made in the abstract and introduction accurately reflect the paper's contributions and scope?
    \item[] Answer: \answerYes{} 
    \item[] Justification: \justificationTODO{}

\item {\bf Limitations}
    \item[] Question: Does the paper discuss the limitations of the work performed by the authors?
    \item[] Answer: \answerYes{} 
    \item[] Justification: In the final section.

\item {\bf Theory Assumptions and Proofs}
    \item[] Question: For each theoretical result, does the paper provide the full set of assumptions and a complete (and correct) proof?
    \item[] Answer: \answerYes{} 
    \item[] Justification: See model section.

    \item {\bf Experimental Result Reproducibility}
    \item[] Question: Does the paper fully disclose all the information needed to reproduce the main experimental results of the paper to the extent that it affects the main claims and/or conclusions of the paper (regardless of whether the code and data are provided or not)?
    \item[] Answer: \answerYes{} 
    \item[] Justification: We will release code.

\item {\bf Open access to data and code}
    \item[] Question: Does the paper provide open access to the data and code, with sufficient instructions to faithfully reproduce the main experimental results, as described in supplemental material?
    \item[] Answer: \answerYes{} 
    \item[] Justification: We will release code.

\item {\bf Experimental Setting/Details}
    \item[] Question: Does the paper specify all the training and test details (e.g., data splits, hyperparameters, how they were chosen, type of optimizer, etc.) necessary to understand the results?
    \item[] Answer: \answerYes{} 
    \item[] Justification: See dataset section.

\item {\bf Experiment Statistical Significance}
    \item[] Question: Does the paper report error bars suitably and correctly defined or other appropriate information about the statistical significance of the experiments?
    \item[] Answer: \answerYes{} 
    \item[] Justification: See experiment section.

\item {\bf Experiments Compute Resources}
    \item[] Question: For each experiment, does the paper provide sufficient information on the computer resources (type of compute workers, memory, time of execution) needed to reproduce the experiments?
    \item[] Answer: \answerYes{} 
    \item[] Justification: See efficiency section.
    
\item {\bf Code Of Ethics}
    \item[] Question: Does the research conducted in the paper conform, in every respect, with the NeurIPS Code of Ethics \url{https://neurips.cc/public/EthicsGuidelines}?
    \item[] Answer: \answerYes{} 

\item {\bf Broader Impacts}
    \item[] Question: Does the paper discuss both potential positive societal impacts and negative societal impacts of the work performed?
    \item[] Answer: \answerYes{} 
    \item[] Justification: See last section.
    
\item {\bf Safeguards}
    \item[] Question: Does the paper describe safeguards that have been put in place for responsible release of data or models that have a high risk for misuse (e.g., pretrained language models, image generators, or scraped datasets)?
    \item[] Answer: \answerYes{} 

\item {\bf Licenses for existing assets}
    \item[] Question: Are the creators or original owners of assets (e.g., code, data, models), used in the paper, properly credited and are the license and terms of use explicitly mentioned and properly respected?
    \item[] Answer:  \answerNA{} 

\item {\bf New Assets}
    \item[] Question: Are new assets introduced in the paper well documented and is the documentation provided alongside the assets?
    \item[] Answer:  \answerNA{} 

\item {\bf Crowdsourcing and Research with Human Subjects}
    \item[] Question: For crowdsourcing experiments and research with human subjects, does the paper include the full text of instructions given to participants and screenshots, if applicable, as well as details about compensation (if any)? 
    \item[] Answer: \answerNA{} 
    \item[] Justification: Not applicable.

\item {\bf Institutional Review Board (IRB) Approvals or Equivalent for Research with Human Subjects}
    \item[] Question: Does the paper describe potential risks incurred by study participants, whether such risks were disclosed to the subjects, and whether Institutional Review Board (IRB) approvals (or an equivalent approval/review based on the requirements of your country or institution) were obtained?
    \item[] Answer: \answerNA{} 
    \item[] Justification: Not applicable.

\end{enumerate}

\end{document}